\documentclass[authoryear]{elsarticle}

\usepackage{lineno}
\modulolinenumbers[5]

\usepackage{hyperref}

\journal{Computerized Medical Imaging and Graphics}

\usepackage{graphicx} 
\usepackage{amsmath}
\usepackage{subfig}
\usepackage{ulem}
\usepackage{color}
\usepackage[table]{xcolor}
\usepackage{setspace}
\usepackage{amssymb}
\usepackage{tabularx}
\usepackage{color,soul} 
\sethlcolor{white} 
\soulregister\ref{7} 
\soulregister\cite{7} 
\soulregister\citep{7} 
\soulregister\citet{7} 
\usepackage[colorinlistoftodos]{todonotes}
\usepackage{bm} 
\usepackage{etoolbox}
\usepackage{siunitx} 
\robustify\bfseries
\usepackage{colortbl}
\usepackage{tabulary}
\usepackage{geometry}
\usepackage{adjustbox} 
\usepackage{rotating}
\colorlet{LightRubineRed}{black!100!}
\colorlet{LightRubineRed1}{red!100!}
\colorlet{LightOrgane}{green!5!orange!95!}
\colorlet{revision}{black!100!}
\definecolor{Mycolor2}{HTML}{00F9DE}
\definecolor{mygray}{gray}{1.0}

\newcommand{\argmax}{\mathop{\rm argmax}\limits}

\begin{document}
	
\begin{frontmatter}

\title{An application of cascaded 3D fully convolutional networks for medical image segmentation}


\tnotetext[t1]{2018. This manuscript version is made available under the CC-BY-NC-ND 4.0 license \url{http://creativecommons.org/licenses/by-nc-nd/4.0/}} 

\author[a]{Holger R. Roth\corref{correspondingauthor}}
\author[a]{Hirohisa Oda}
\author[b]{Xiangrong Zhou}
\author[a]{Natsuki Shimizu}
\author[a]{Ying Yang}
\author[a]{Yuichiro  Hayashi}
\author[a]{Masahiro Oda}
\author[c]{Michitaka Fujiwara}
\author[d]{Kazunari Misawa}
\author[a]{Kensaku Mori\corref{correspondingauthor}}

\cortext[correspondingauthor]{\href{mailto:rothhr@mori.m.is.nagoya-u.ac.jp}{rothhr@mori.m.is.nagoya-u.ac.jp} or \href{mailto:kensaku@is.nagoya-u.ac.jp}{kensaku@is.nagoya-u.ac.jp}}

\address[a]{Nagoya University, Furo-cho, Chikusa-ku, Nagoya, Japan}
\address[b]{Gifu University, Yanagido, Gifu, Japan}
\address[c]{Nagoya University Graduate School of Medicine, Nagoya, Japan}
\address[d]{Aichi Cancer Center, Kanokoden, Chikusa-ku, Nagoya, Japan}

\begin{abstract}
Recent advances in 3D fully convolutional networks (FCN) have made it feasible to produce dense voxel-wise predictions of volumetric images. In this work, we show that a multi-class 3D FCN trained on manually labeled CT scans of several anatomical structures (ranging from the large organs to thin vessels) can achieve competitive segmentation results, while avoiding the need for handcrafting features or training class-specific models. 	
	
To this end, we propose a two-stage, coarse-to-fine approach that will first use a 3D FCN to roughly define a candidate region, which will then be used as input to a second 3D FCN. This reduces the number of voxels the second FCN has to classify to $\sim$10\% and allows it to focus on more detailed segmentation of the organs and vessels. 
	
We utilize training and validation sets consisting of 331 clinical CT images and test our models on a completely unseen data collection acquired at a different hospital that includes 150 CT scans, targeting three anatomical organs (liver, spleen, and pancreas). In challenging organs such as the pancreas, our cascaded approach improves the mean Dice score from 68.5 to 82.2\%, achieving the highest reported average score on this dataset. We compare with a 2D FCN method on a separate dataset of 240 CT scans with 18 classes and achieve a significantly higher performance in small organs and vessels. Furthermore, we explore fine-tuning our models to different datasets.
	
Our experiments illustrate the promise and robustness of current 3D FCN based semantic segmentation of medical images, achieving state-of-the-art results.\footnote{Our code and trained models are available for download: \url{github.com/holgerroth/3Dunet_abdomen_cascade}}.	
\end{abstract}


\end{frontmatter}


\clearpage
\newpage
\section{Introduction} 
Recent advances in fully convolutional networks (FCN) have made it feasible to train models for pixel-wise segmentation in an end-to-end fashion \citep{long2015fully}. Efficient implementations of 3D convolution and growing GPU memory have made it possible to extent these methods to 3D medical imaging and train networks on large amounts of annotated volumes. One such example is the recently proposed 3D U-Net \citep{cciccek20163d}, which applies a 3D FCN with skip connections to sparsely annotated biomedical images. Alternative architectures for processing volumetric images have also been successfully applied to 3D medical image segmentation \citep{milletari2016v,chen2016voxresnet,dou20173d}.
In this work, we show that a 3D FCN, like 3D U-Net, trained on manually labeled data of several anatomical structures (ranging from the large organs to thin vessels) can also achieve competitive segmentation results on clinical CT images, very different from the original application of 3D U-Net using confocal microscopy images. We furthermore compare our approach to 2D FCNs applied to the same images. 

Our approach applies 3D FCN architectures to problems of multi-organ and vessel segmentation in a cascaded fashion. A FCN can be trained on whole 3D CT scans. However, because of the high imbalance between background and foreground voxels (organs, vessels, etc.) the network will concentrate on differentiating the foreground from the background voxels in order to minimize the loss function used for training. While this enables the FCN to roughly segment the organs, it causes particularly smaller organs (like the pancreas or gallbladder) and vessels to suffer from inaccuracies around their boundaries. 

To overcome this limitation, we learn a second-stage FCN  in a cascaded manner that focuses more on the boundary regions. This is a coarse-to-fine approach in which the first-stage FCN sees around 40\% of the voxels using only a simple automatically generated mask of the patient's body. In the second stage, the amount of the image\rq{}s voxels is further reduced to around 10\%. In effect, this step narrows down and simplifies the search space for the FCN to decide which voxels belong to the background or any of the foreground classes; this strategy has been successful in many computer vision problems \citep{viola2004robust,li2016iterative}. Our approach is illustrated on a training example in Fig. \ref{fig:coarse-to-fine}. 
\subsection{Related work}
Multi-organ segmentation has attracted considerable interest over the years. Classical approaches include statistical shape models \citep{cerrolaza2015automatic,okada2015abdominal}, and/or employ techniques based on image registration. So called multi-atlas label fusion \citep{rohlfing2004evaluation,wang2013multi,iglesias2015multi} has found wide application in clinical research and practice. Approaches that combine techniques from multi-atlas registration and machine learning are also common place and have been successfully applied to multi-organ segmentation in abdominal imaging \citep{tong2015discriminative,oda2016regression}.  However, a fundamental disadvantage of image registration based methods is there extensive computational cost \citep{iglesias2015multi}. Typical methods need hours of computation time in order to complete on single desktop machines \citep{wolz2013automated}.

The recent success of deep learning based classification and segmentation methods are now transitioning to applications of multi-class segmentation in medical imaging. Recent examples of deep learning applied to organ segmentation include \citep{roth2017spatial,zhou2016pancreas,christ2016automatic,zhou2016three}. Many methods focus on the segmentation of single organs like prostate \citep{milletari2016v}, liver \citep{christ2016automatic}, or pancreas \citep{roth2015deeporgan,roth2016spatial}. Multi-organ segmentation in abdominal CT has also been approached by works like \citep{hu2017automatic,gibson2017towards}. Most methods are based on variants of FCNs \citep{long2015fully} that either employ 2D convolutional layers in a slice-by-slice fashion \cite{roth2016spatial,zhou2016pancreas,christ2016automatic,zhou2016three}, 2D convolutions on orthogonal (2.5D) cross-sections  \citep{roth2015deeporgan,prasoon2013deep}, and 3D convolutional layers \citep{milletari2016v,chen2016voxresnet,dou20173d,kamnitsas2017efficient}. A common feature of these novel segmentation methods is that they are able to extract the features useful for image segmentation directly from the training imaging data, which is crucial for the success of deep learning \citep{lecun2015deep}. This avoids the need for \textit{hand-crafting} features that are suitable for detection of individual organs.
\subsection{Contributions}
Due to the automatic learning of image feature and in contrast to previous approaches of multi-organ segmentation where separate models have to be created for each organ \citep{oda2016regression,tong2015discriminative}, our proposed method allows us to use the same model to segment very different anatomical structures such as large abdominal organs (liver, spleen), but also vessels like arteries and veins. Furthermore, other recent FCN-based methods that applied in medical imaging in cascaded/iterative fashion were often constrained to using rectangular bounding boxes around single organs \citep{roth2017spatial,zhou2016pancreas} and/or performing slice-wise processing in 2D \citep{christ2016automatic,zhou2016three}.
\begin{figure}[htb]
	\centering
	\includegraphics[width=0.95\linewidth, clip]{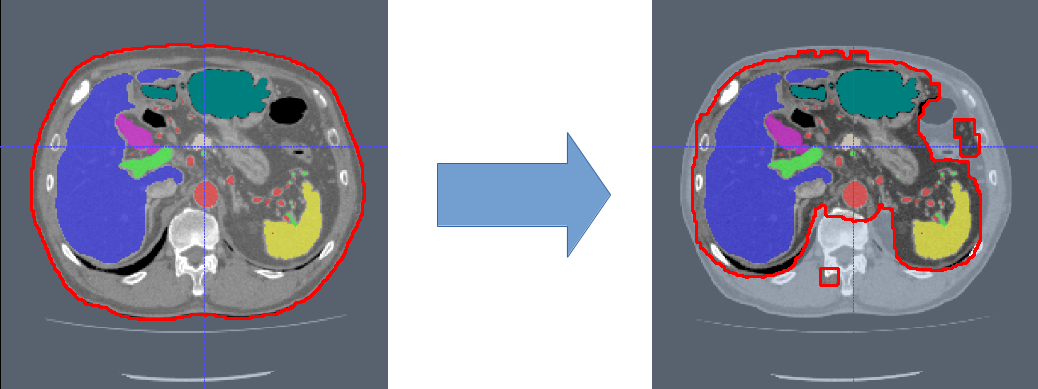}
	\caption{Cascaded 3D fully convolutional networks in a coarse-to-fine approach: the first stage (left) learns the generation of a candidate region for training a second-stage FCN (right) for finer prediction.  Outlined red area shows candidate region $C_1$ used in first stage and $C_2$ used in second stage. Colored regions denote ground truth annotations for training (best viewed in color).}
	\label{fig:coarse-to-fine}
\end{figure}
\section{Methods}
\begin{figure*}[htb]
	\centering
	\adjincludegraphics[width=1.0\textwidth]{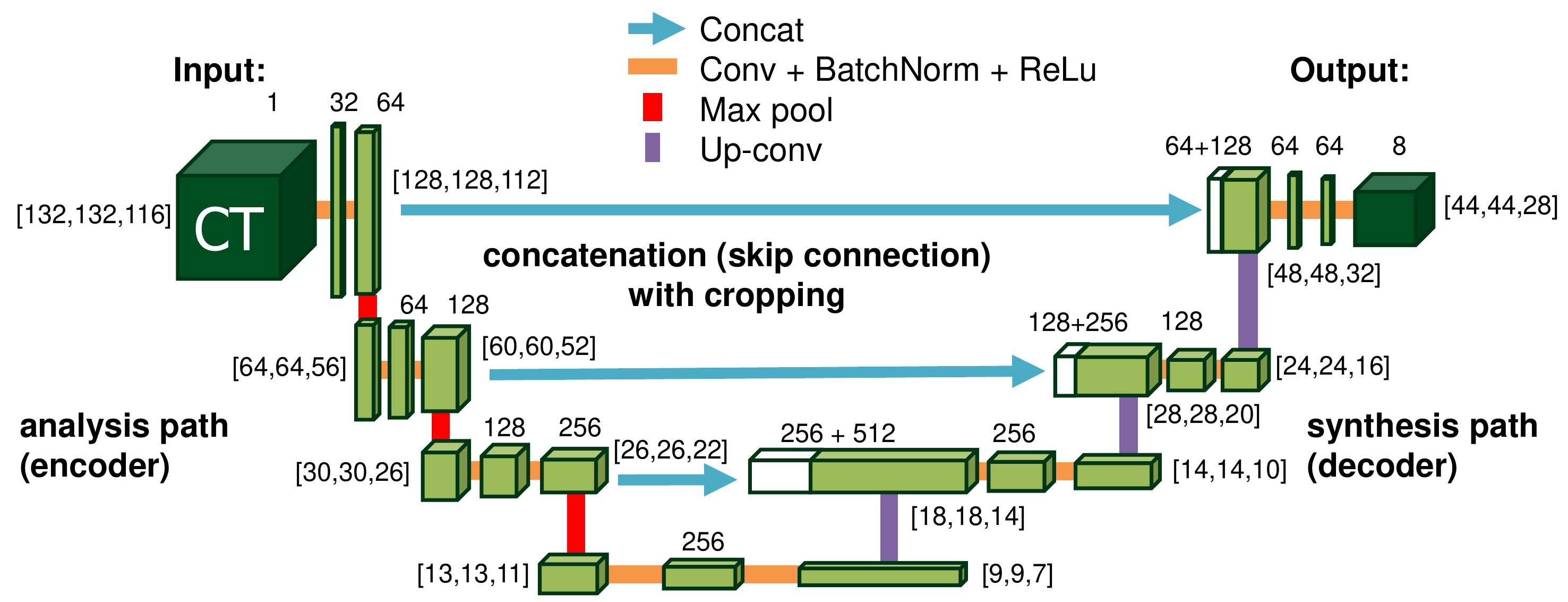}
	\caption{\textcolor{revision}{The architecture of 3D U-Net \citep{cciccek20163d}, a type of fully convolutional network. It applies an end-to-end architecture using only valid convolutions (\textit{Conv}) with no padding and kernel sizes of $3\times3\times3$. Rectified Linear units (\textit{ReLU}) are used as activation functions. This results in a smaller output size than input size and requires cropping of when mapping lower level feature maps of the analysis path to the synthesis path of the network via concatenation (\textit{Concat}). Max-pooling (\textit{Max pool}) is used to reduce the resolution of feature maps, while up-convolutions (\textit{Up-conv}) are used for up-sampling the feature maps back to higher resolutions. The number of extracted feature maps is noted above each layer. We show the input and output size of feature maps at each level of the network. These parameters are kept constant for all experiments performed in this study. Batch normalization (\textit{BatchNorm}) is used throughout the network for improved convergence \citep{ioffe2015batch}.} \label{fig:3Dunet}}
\end{figure*}

Convolutional neural networks have the ability to solve challenging classification tasks in a data-driven manner. Given a training set of images and labels $\mathbf{S} = \left\{(I_n,L_n), n = 1,\dots,N\right\}$, $I_n$ denotes the raw CT images and $L_n$ denotes the ground truth label images. Each $L_n$ contains $K$ class labels consisting of the manual segmentations of the foreground anatomy (e.g. artery, portal vein, lungs, liver, spleen, stomach, gallbladder, and pancreas) and the background for each voxel in the CT image. Our employed network architecture is the 3D extension by \citet{cciccek20163d} of the U-Net proposed by \citet{ronneberger2015u}. U-Net, which is a type of fully convolutional network (FCN) \citep{long2015fully} was originally proposed for bio-medical image applications, utilizes deconvolution \citep{long2015fully} (or sometimes called up-convolutions \citep{cciccek20163d}) to remap the lower resolution feature maps within the network to the denser space of the input images. This operation allows for denser voxel-to-voxel predictions in contrast to previously proposed sliding-window CNN methods where each voxel under the window is classified independently making such architecture inefficient for processing large 3D volumes. In 3D U-Net, operations such as 2D convolution, 2D max-pooling, and 2D deconvolution are replaced by their 3D counterparts \citep{cciccek20163d}. We use the open-source implementation of 3D U-Net\footnote{\url{http://lmb.informatik.uni-freiburg.de/resources/opensource/unet.en.html}} based on the Caffe deep learning library \citep{jia2014caffe}. The 3D U-Net architecture consists of analysis and synthesis paths with four resolution levels each. Each resolution level in the analysis path contains two $3 \times 3 \times 3$ convolutional layers, each followed by rectified linear units (ReLU) and a $2 \times 2 \times 2$ max pooling with strides of two in each dimension. In the synthesis path, the convolutional layers are replaced by deconvolutions of $2 \times 2 \times 2$ with strides of two in each dimension. These are followed by two $3 \times 3 \times 3$ convolutions, each of which has a ReLU. Furthermore, 3D U-Net employs shortcut (or skip) connections from layers of equal resolution in the analysis path to provide higher-resolution features to the synthesis path \citep{cciccek20163d}. The last layer contains a $1\times 1\times 1$ convolution that reduces the number of output channels to the number of class labels $K$. This architecture has over 19 million learnable parameters and can be trained to minimize a weighted voxel-wise cross-entropy loss \citep{cciccek20163d}. A schematic illustration of 3D U-Net is shown in Fig. \ref{fig:3Dunet}.
\subsection{Loss function: adjustments for multi-organ segmentation}
\textcolor{revision}{The voxel-wise cross-entropy loss is defined as} 
\begin{equation} \small     
	\mathcal{L} \ = \ \frac{-1}{N}\sum^{K}_{k=1} \left( \sum_{x\in S_k}\log{\left(\hat{p}_{k}(x)\right)} \right),
\label{equ:loss}
\end{equation}
\textcolor{revision}{where $\hat{p}_{k}$ are the \textit{softmax} class probabilities}
\begin{equation} \small
	\hat{p}_{k}(x) \ = \ \frac{\exp(x_{k}(x))}{    \sum^{K}_{k\rq{}=1}\exp(x_{k\rq{}}(x))       },
\end{equation} 		
\textcolor{revision}{$N$ are the total number of voxels $x$, $S_k$ is the set of voxels within one class in $L_n$, and $k \in [1,2,\dots,K]$ indicates the ground truth class label. The input to this loss function is real valued output predictions $x \in [-\infty,+\infty]$ from the last convolutional layer.} 

\textcolor{revision}{However, in most cases minimizing this loss will instantly make the network converge to classifying every voxel as background. This is because of the large dominance of the background class in the images. In order to combat this large data imbalance between foreground/background voxels and differently sized organs and vessels, we apply a voxel-wise weight $\lambda_k$ to this loss function (Eq. \ref{equ:loss}). In this work, we choose $\lambda_i$ such that $\sum^{K}_{k=1} \lambda_k = 1$, with}
\begin{equation} 
	\lambda_k = \frac{1-N_k/N_C}{K-1},
	\label{equ:weight}
\end{equation}
\textcolor{revision}{where $N_k$ is the number of voxels in each class $S_k$, and $N_C$ is the number of voxels within a candidate region $C_1$ or $C_2$. The weights $\lambda_i$ help to balance the common voxels (i.e., background) with respect to such smaller organs as vessels or the pancreas by giving more weight to the latter.}

\textcolor{revision}{Now, the weighted cross-entropy loss can be written as:}
\begin{equation} \small
	\mathcal{L} \ = \ \frac{-1}{N}\sum^{K}_{k=1} \lambda_k \left( \sum_{x\in S_k}\log{\left(\hat{p}_{k}(x)\right)} \right),
	\label{equ:wloss}
\end{equation}
\textcolor{revision}{We use the loss formulation in Eq. \ref{equ:wloss} for all experiments in this paper.}
\subsection{Coarse-to-fine prediction}
In our experiments, the input to the network is fixed to a given size $N_\mathrm{x}\times N_\mathrm{y}\times N_\mathrm{z}$, mainly influenced by considerations of available memory size on the GPU. In training, sub-volumes of that given size are randomly sampled from the candidate regions within the training CT images, as described below. To increase the field of view presented to the CNN and reduce informative redundancy among neighboring voxels, each image is downsampled by a factor of 2. The resulting prediction maps are then resampled back to the original resolution using nearest neighbor interpolation (or linear interpolation in case of the probability maps). 

\paragraph{\textcolor{revision}{1\textsuperscript{st} Stage}}
In the first stage, we apply simple thresholding in combination with morphological operations (hole filling and largest component selection) to get a mask of the patient's body. This mask can be utilized as candidate region $C_1$ to reduce the number of voxels necessary to compute the network's loss function and reduce the amount of input 3D regions shown to the CNN during training to about 40\%. 
\paragraph{\textcolor{revision}{2\textsuperscript{nd} Stage}}
\textcolor{revision}{After training the first-stage FCN}, it is applied to each image to generate candidate regions $C_2$ for training the second-stage FCN \textcolor{revision}{(see Fig. \ref{fig:coarse-to-fine})}. We define the \textcolor{revision}{predicted} organ labels in the testing phase using the $\argmax$ of the class probability maps. \textcolor{revision}{All foreground labels are then dilated in 3D using a voxel radius of $r$ in order to compute $C_2$, resulting in a binary candidate map}. 

\begin{figure}[htb]
	\centering
	\subfloat[Training]{\includegraphics[width=0.45\textwidth]{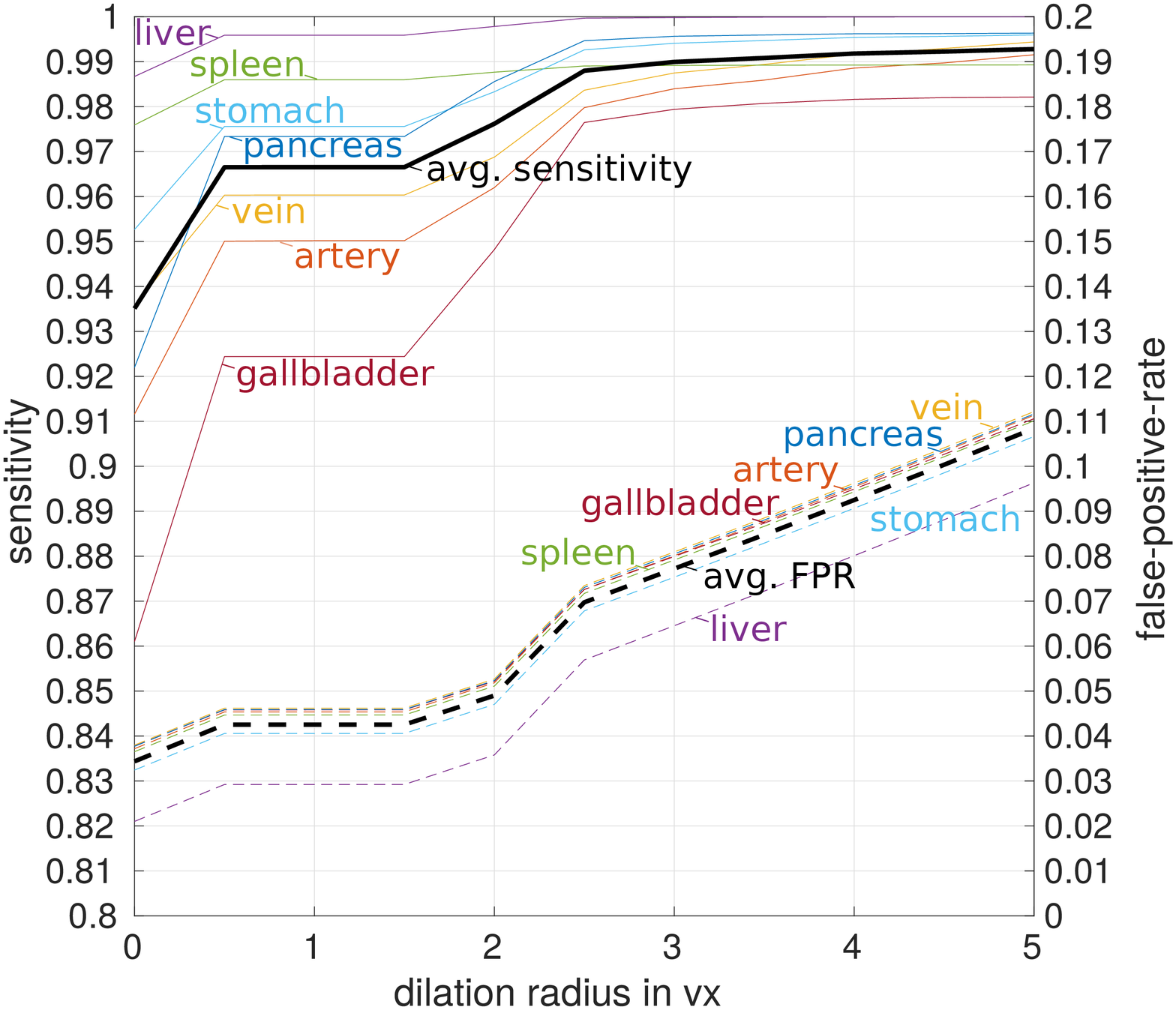}}
	\hfill
	\subfloat[Validation]{\includegraphics[width=0.45\textwidth]{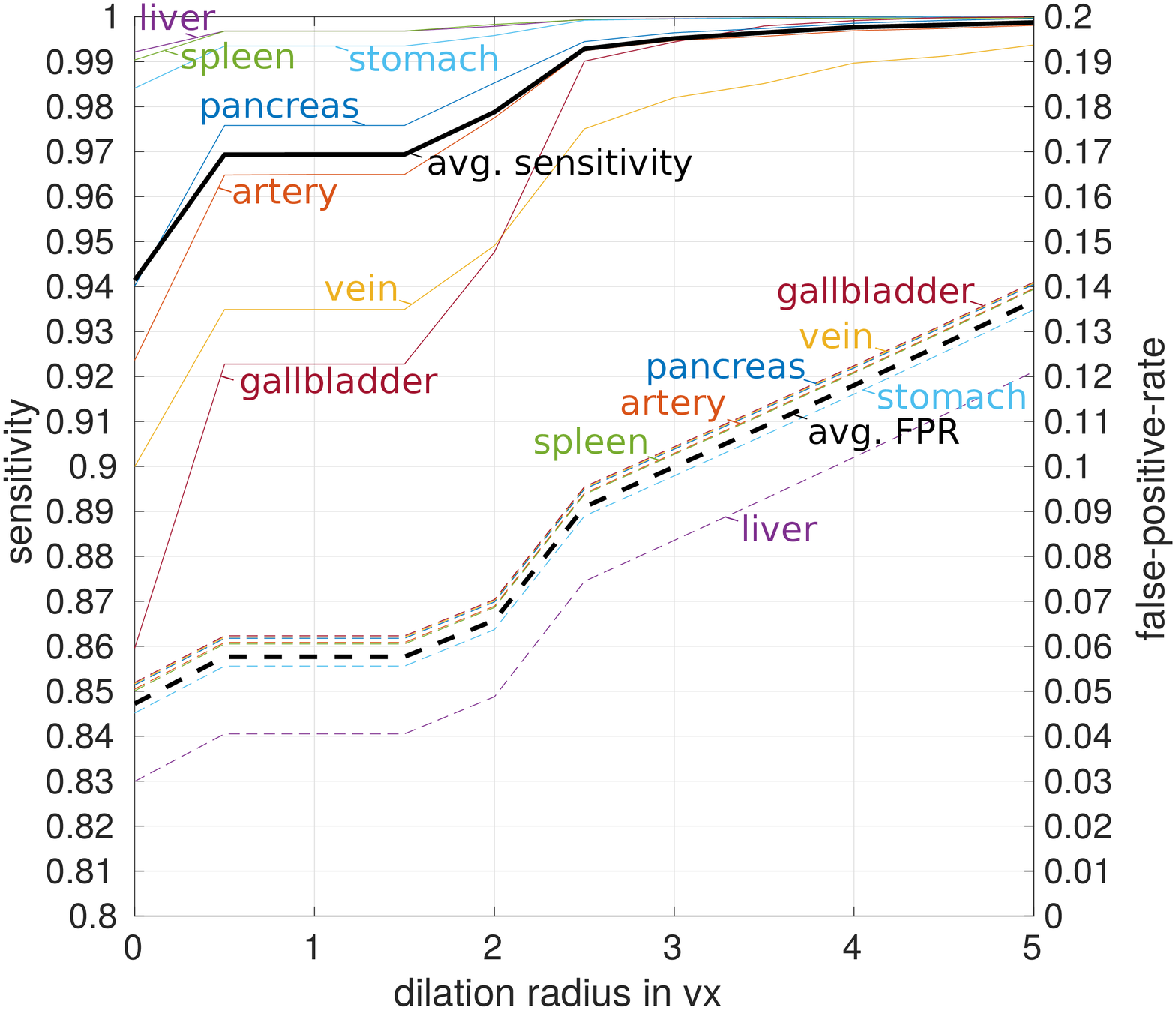}}    
	\caption{Sensitivity and false-positive-rate (FPR) as a function of dilating prediction maps of first stage in training (a) and validation (b). We observe good trade-off between high sensitivity ($>$99\% on average) and low false-positive-rate ($\sim$10\% on average) at dilation radius of $r=3$.}
	\label{fig:recall_stage1}
\end{figure}

When comparing the recall and false-positive rates of \textcolor{revision}{the} first-stage FCN with respect to $r$ for both the training and validation sets, $r=3$ gives good trade-off between high recall ($>$99\%) and low false-positive rates ($\sim$10\%) for each organ on our training and validation sets (see Fig. \ref{fig:recall_stage1}). 

\begin{figure*}[htb] 
	\centering
	\adjincludegraphics[width=0.95\textwidth]{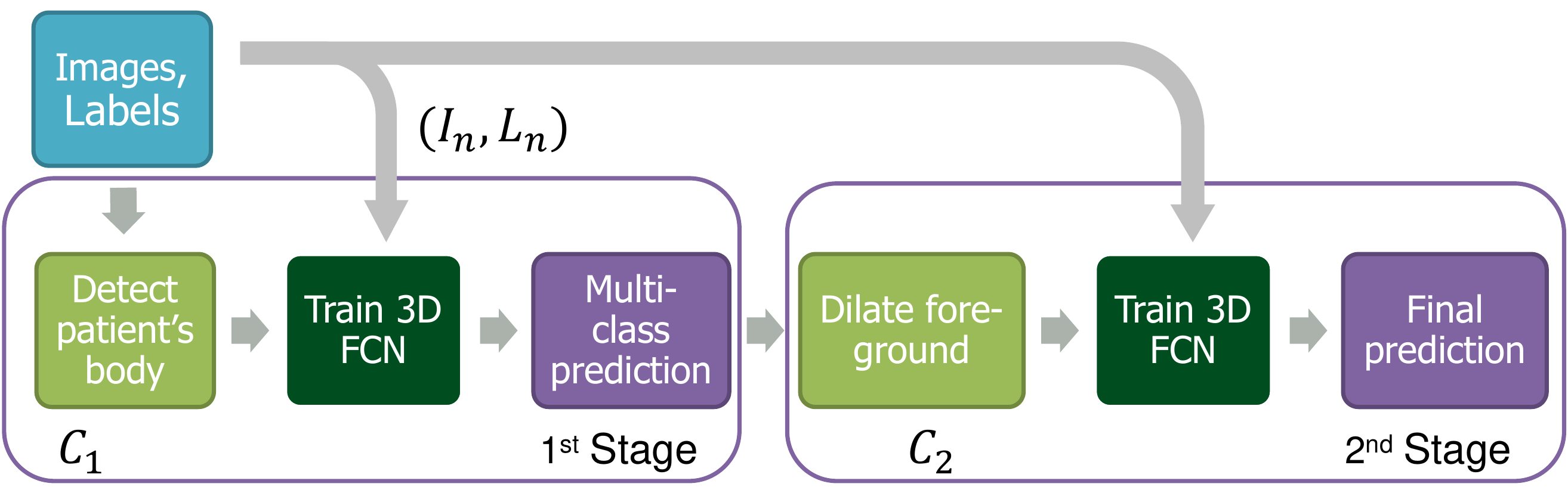}
	\caption{\textcolor{revision}{Flowchart of our multi-stage cascaded training scheme.}}
	\label{fig:flowchart}
\end{figure*}

\textcolor{revision}{Our overall multi-stage training scheme is illustrated in Fig. \ref{fig:flowchart}}
\subsection{Training}
\begin{figure}[htb]
	\centering
	\subfloat{\includegraphics[width=0.24\textwidth]{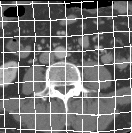}}
	\hfill
	\subfloat{\includegraphics[width=0.24\textwidth]{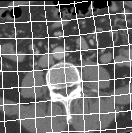}}
	\hfill
	\subfloat{\includegraphics[width=0.24\textwidth]{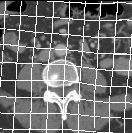}}
	\hfill
	\subfloat{\includegraphics[width=0.24\textwidth]{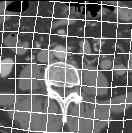}}			
	\caption{Axial cross-section through the same patient CT image at various examples of plausible random deformation during training. A deformed grid pattern is overlaid in order to better illustrate the applied deformation. At each iteration, the random deformation is computed on the fly.}
	\label{fig:deformations}
\end{figure}

The network iteratively adjusts its parameters by stochastic gradient descent. Batch normalization is used throughout the network for improved convergence and we utilize random elastic deformations in 3D during training to artificially increase the amount of available data samples and increase robustness, similar to \citep{cciccek20163d}. Hence, we randomly sample deformation fields from a uniform distribution with a maximum
displacement of $\pm$4 and a grid spacing of 32 voxels (see Fig. \ref{fig:deformations}). Furthermore, we applied random rotations between $-5^{\circ}$ and $+5^{\circ}$, and translations of -20 to 20 voxels in each direction at each iteration in order to generate plausible deformations during training. Each training sub-volume is randomly extracted from $C_1$ or $C_2$ in both stages. 
\subsection{Testing}
\begin{figure}[htb] 
	\centering
	\includegraphics[width=0.85\textwidth]{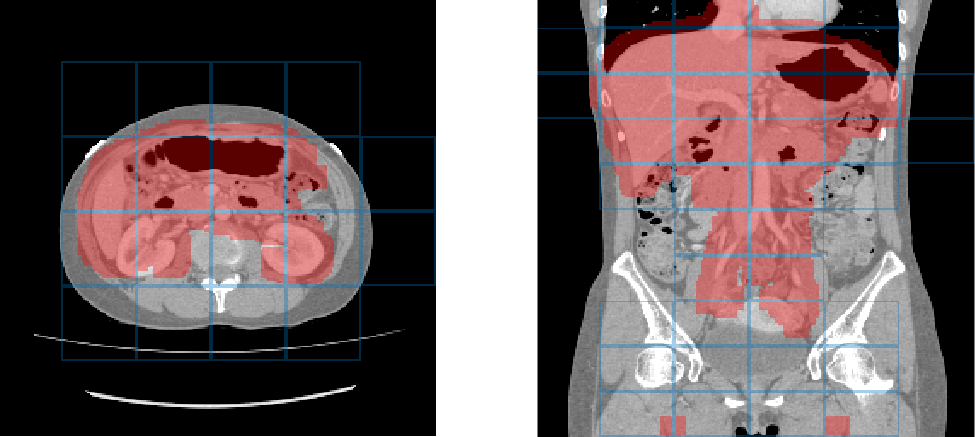}
	\caption{The non-overlapping tiling approach on second stage candidate region $C_2$. Note that the grid shows the output tiles of size $44 \times 44 \times 28$ ($x,y,z$-directions). Each predicted tile is based on a larger input of $132 \times 132 \times 116$ that the network processes.}
	\label{fig:tiling}
\end{figure}

The CT image is processed by the 3D FCN using a tiling strategy (sliding-window) \citep{cciccek20163d} as illustrated in Fig. \ref{fig:tiling}. For greater speed, we use non-overlapping tiles in the first stage and investigate the use of non-overlapping and overlapping tiles in the second. When using overlapping tiles (with a $4 \times$ higher sampling rate of each voxel $x$), the resulting probabilities for the overlapping voxels are averaged: 
\begin{equation} \small
p(x) = \frac{1}{R}\sum^{R}_{r=1} \ p_r(x).
\label{equ:overlapping}
\end{equation}
\section{Experiments \& Results}
\begin{figure}[htb] 
	\centering
	\subfloat[Ground truth]{\includegraphics[width=0.35\textwidth]{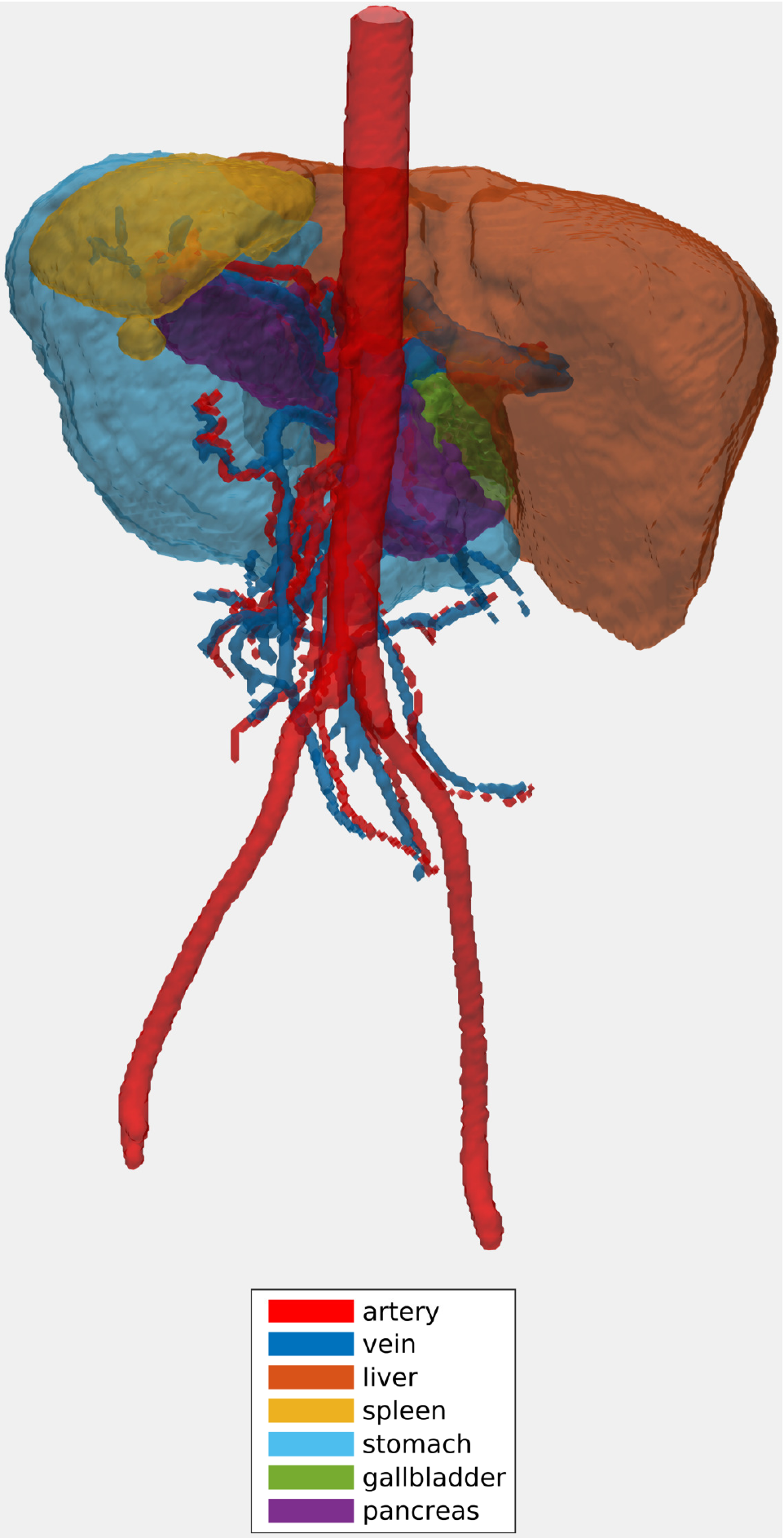}}
	\hspace{5em}
	\subfloat[Stage 2 - N/OL]{\includegraphics[width=0.35\textwidth]{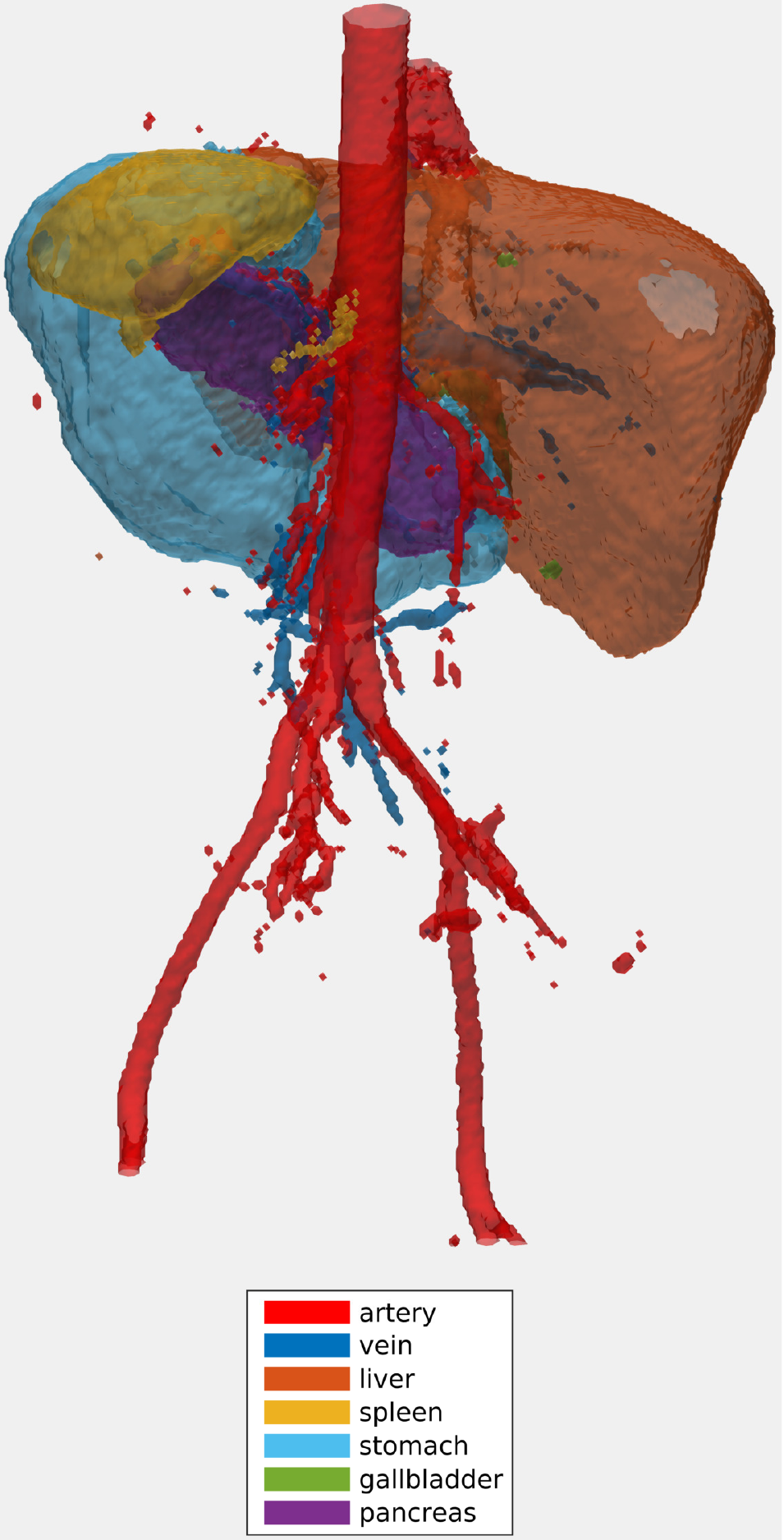}}    
	\caption{Example of the validation set with (a) ground truth and (b) the corresponding non-overlapping (N/OL) segmentation result. The posterior to anterior view is shown to visualize the inner organs.}
	\label{fig:prediction_rendering}
\end{figure}
\subsection{Training and validation} 
\label{sec:train-and-val}
Our dataset includes 331 contrast-enhanced abdominal clinical CT images in the portal venous phase used for pre-operative planning in gastric surgery. Each CT volume consists of $460-1177$ slices of $512\times 512$ pixels. The voxel dimensions are [0.59-0.98, 0.59-0.98, 0.5-1.0] mm. A random split of 281/50 patients is used for training and validating the network, i.e., determining when to stop training to avoid overfitting. In order to generate plausible deformations during training, we sample from a normal distribution with a standard derivation of 4 and a grid spacing of 32 voxels, and apply random rotations between $-5^\circ$ and $+5^\circ$ to the training images. No deformations were applied during testing. We trained 200,000 iterations in the first stage and 115,000 in the second. Table \ref{tab:dice_results_validation} summarizes the Dice similarity scores for each organ labeled in the 50 validation cases. On average, we achieved a 7.5\% improvement in Dice scores per organ. Small, thin organs such as arteries especially benefit from our two-stage cascaded approach. For example, the mean Dice score for arteries improved from 59.0 to 79.6\% and from 54.8 to 63.1\% for the pancreas. The effect is less pronounced for large organs, like the liver, the spleen, and the stomach. Fig. \ref{fig:prediction_rendering} shows an example result from the validation set and illustrates the tiling approach. The 3D U-Net separates the foreground organs well from the background tissue of the images. 
\begin{table}
	\tiny
	\centering
	\caption{\textbf{Validation set:} Dice similarity score [\%] of different stages of FCN processing}
	\label{tab:dice_results_validation}
	\begin{tabular}{lrrrrrrrr}
		\multicolumn{8}{l}{\textbf{Stage 1: Non-overlapping}} \tabularnewline
		\hline
		\rowcolor[gray]{.9}\textbf{Dice} & \textbf{artery} & \textbf{\ \ \ \ vein} & \textbf{\ \ \ liver} & \textbf{\ \ spleen} & \textbf{\ stomach} & \textbf{\ gallbladder} & \textbf{\ pancreas} & \textbf{ \ \ Mean}\tabularnewline
		\hline
		\textbf{Mean} & 59.0 & 64.7 & 89.6 & 84.1 & 80.0 & 69.6 & 54.8 & 71.7\tabularnewline
		\rowcolor[gray]{.9}\textbf{Std} & 7.8 & 8.6 & 1.7 & 4.7 & 18.3 & 14.1 & 11.0 & 9.5\tabularnewline
		\textbf{Median} & 59.8 & 67.3 & 90.0 & 85.2 & 87.5 & 73.2 & 57.2 & 74.3\tabularnewline
		\rowcolor[gray]{.9}\textbf{Min} & 41.0 & 34.5 & 84.4 & 70.9 & 8.4 & 13.8 & 23.5 & 39.5\tabularnewline
		\textbf{Max} & 75.7 & 76.0 & 92.6 & 91.4 & 94.8 & 86.8 & 72.0 & 84.2\tabularnewline
		\hline
		&  &  &  &  &  &  &  \tabularnewline
		\multicolumn{8}{l}{\textbf{Stage 2: Non-overlapping}} \tabularnewline
		\hline
		\rowcolor[gray]{.9}\textbf{Dice} & \textbf{artery} & \textbf{vein} & \textbf{liver} & \textbf{spleen} & \textbf{stomach} & \textbf{gallbladder} & \textbf{pancreas} & \textbf{Mean}\tabularnewline
		\hline
		\textbf{Mean} & 79.6 & 73.1 & 93.2 & 90.6 & 84.3 & 70.6 & 63.1 & 79.2\tabularnewline
		\rowcolor[gray]{.9}\textbf{Std} & 6.5 & 7.9 & 1.5 & 2.8 & 17.3 & 15.9 & 10.7 & 8.9\tabularnewline
		\textbf{Median} & 82.3 & 74.6 & 93.5 & 91.2 & 90.9 & 77.3 & 64.5 & 82.1\tabularnewline
		\rowcolor[gray]{.9}\textbf{Min} & 62.9 & 33.3 & 88.9 & 82.3 & 10.9 & 13.0 & 32.4 & 46.2\tabularnewline
		\textbf{Max} & 87.0 & 83.2 & 95.6 & 95.1 & 96.3 & 89.4 & 81.8 & 89.8\tabularnewline
		\hline
		&  &  &  &  &  &  &  \tabularnewline
		\multicolumn{8}{l}{\textbf{Stage2 vs Stage1}} \tabularnewline
		\hline
		\rowcolor[gray]{.9}\textbf{Dice} & \textbf{artery} & \textbf{vein} & \textbf{liver} & \textbf{spleen} & \textbf{stomach} & \textbf{gallbladder} & \textbf{pancreas} & \textbf{Mean}
		\tabularnewline
		\hline
		\textbf{Mean} & 20.61 & 8.41 & 3.60 & 6.42 & 4.22 & 0.93 & 8.26 & 7.49\tabularnewline
		\rowcolor[gray]{.9}\textbf{Std} & -1.24 & -0.68 & -0.18 & -1.97 & -0.97 & 1.78 & -0.35 & -0.52\tabularnewline
		\textbf{Median} & 22.57 & 7.34 & 3.42 & 6.00 & 3.44 & 4.15 & 7.31 & 7.75\tabularnewline
		\rowcolor[gray]{.9}\textbf{Min} & 21.83 & -1.20 & 4.47 & 11.35 & 2.44 & -0.75 & 8.85 & 6.71\tabularnewline
		\textbf{Max} & 11.28 & 7.21 & 3.06 & 3.70 & 1.52 & 2.67 & 9.74 & 5.60\tabularnewline
	\end{tabular}
\end{table}
\subsection{Testing} 
\begin{figure}[htb] 
	\centering
	\subfloat[Ground truth]{\includegraphics[width=0.235\textwidth]{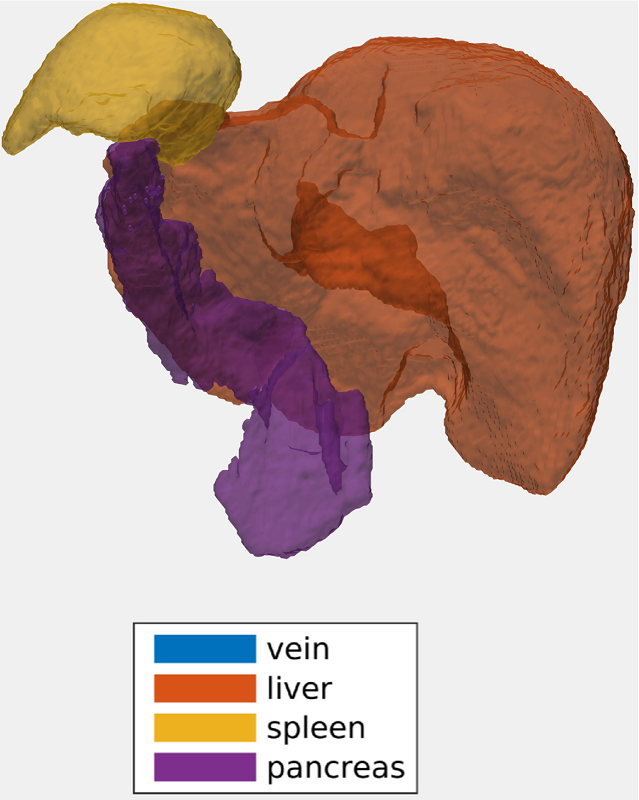}}
	\hfill
	\subfloat[Stage 1]{\includegraphics[width=0.24\textwidth]{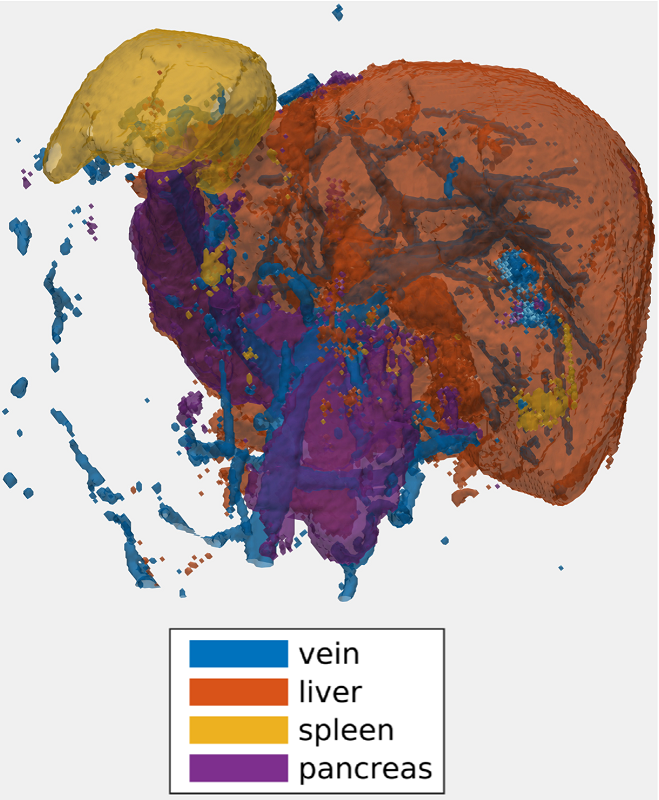}}  
	\hfill
	\subfloat[Stage 2: N/OL]{\includegraphics[width=0.24\textwidth]{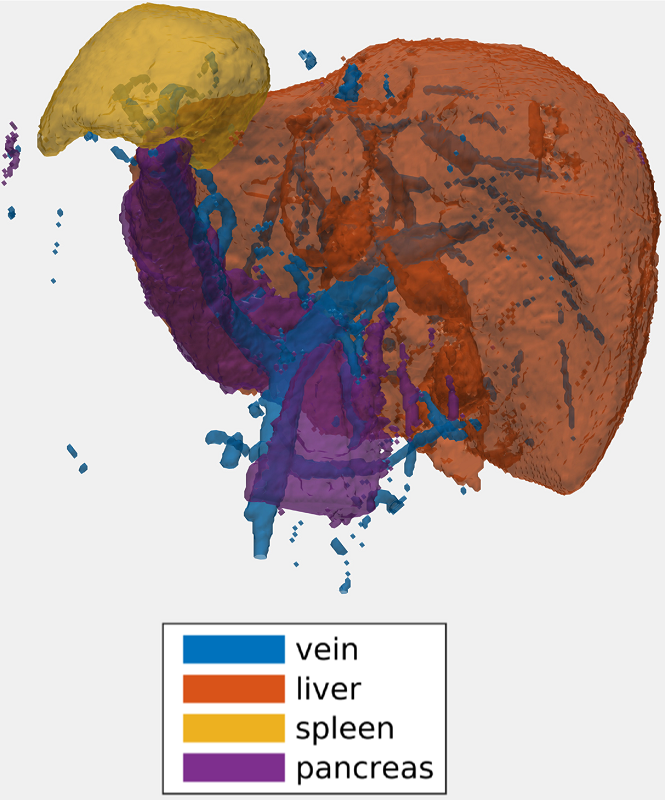}}    
	\hfill
	\subfloat[Stage 2: OL]{\includegraphics[width=0.24\textwidth]{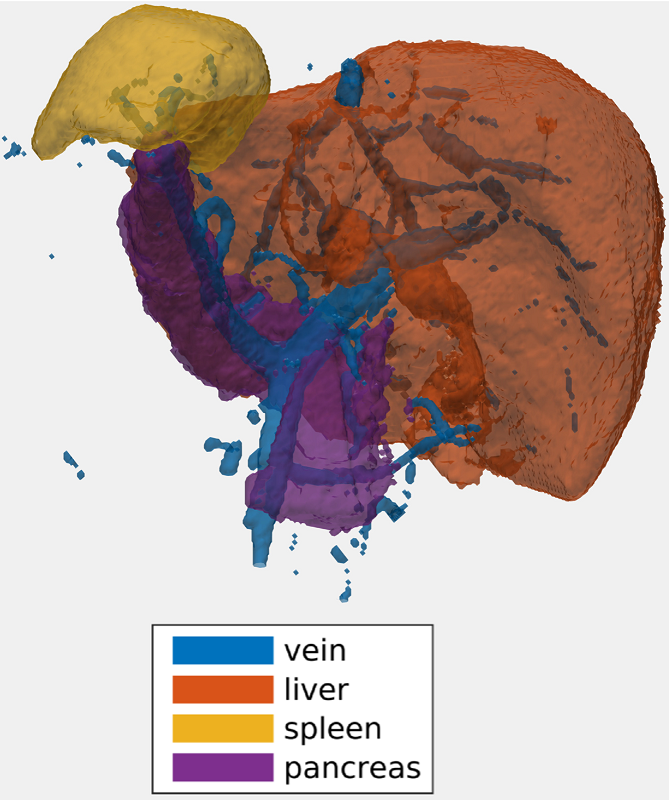}}        
	\caption{Surface renderings: 
		(a) ground truth segmentation, 
		(b) result of proposed method in first stage, second-stage results using (c) non-overlapping (N/OL), and (d) overlapping (OL) tiles strategy. The posterior to anterior view is shown for better visualization of the pancreas.}
	\label{fig:surface_stages}
\end{figure}

Our test set is different from our training and validation data. It originates from a different hospital, scanners, and research study with gastric cancer patients. 150 abdominal CT scans were acquired in the portal venous phase. Each CT volume consists of $263-1061$ slices of $512\times 512$ pixels. Voxel dimensions are [0.55-0.82, 0.55-0.82, 0.4-0.80] mm. The pancreas, liver, and spleen were semi-automatically delineated by three trained researchers and confirmed by a clinician. Figure \ref{fig:surface_stages} shows surface renderings for comparison of the different stages of the algorithm. A typical testing case in the first and second stages is shown using non-overlapping and overlapping tiles. Dice similarity scores are listed in Table \ref{tab:dice_results_testing}. The second stage achieves the highest reported average score for pancreas in this dataset with 82.2\% $\pm$10.2\%. Previous state of the art on this dataset was at 75.1\% $\pm$15.4\% while using leave-one-out-validation \citep{oda2016regression}.

The testing dataset provides slightly higher image quality than our training/validation dataset. Furthermore, its field of view is more constrained to the upper abdomen. This likely explains the improved performance for liver and pancreas compared to the validation set in Table \ref{tab:dice_results_validation}.
\clearpage
\newpage
\begin{table}
	\small
	\centering
	\caption{\textbf{Testing on unseen dataset:} Dice similarity score [\%] of different stages of FCN processing.}
	\label{tab:dice_results_testing}
	\begin{tabular}{l|rrr|lrrr|lrrr}
		& \multicolumn{3}{l}{\textbf{Stage 1: Non-overlapping}} &\ \ \ & \multicolumn{3}{l}{\textbf{Stage 2: non-overlapping}} &\ \ \ & \multicolumn{3}{l}{\textbf{Stage 2: Overlapping}} \tabularnewline
		\hline
		\rowcolor[gray]{.9}\textbf{Dice} & \textbf{liver} & \textbf{spleen} & \textbf{pancreas} & \ & \textbf{liver} & \textbf{spleen} & \textbf{pancreas} & \ & \textbf{liver} & \textbf{spleen} & \textbf{pancreas}\tabularnewline
		\hline
		\textbf{Mean} & 93.6 & 89.7 & 68.5 &  & 94.9 & 91.4 & 81.2 &  & 95.4 & 92.8 & 82.2\tabularnewline
		\rowcolor[gray]{.9}\textbf{Std} & 2.5 & 8.2 & 8.2 &  & 2.1 & 8.9 & 10.2 &  & 2.0 & 8.0 & 10.2\tabularnewline
		\textbf{Median} & 94.2 & 91.8 & 70.3 &  & 95.4 & 94.2 & 83.1 &  & 96.0 & 95.4 & 84.5\tabularnewline
		\rowcolor[gray]{.9}\textbf{Min} & 78.2 & 20.6 & 32.0 &  & 80.4 & 22.3 & 1.9 &  & 80.9 & 21.7 & 1.8\tabularnewline
		\textbf{Max} & 96.8 & 95.7 & 82.3 &  & 97.3 & 97.4 & 91.3 &  & 97.7 & 98.1 & 92.2\tabularnewline
		\hline
	\end{tabular}
\end{table}
\clearpage
\newpage
\subsection{Comparison to other methods} 
Even though direct comparison is difficult due to the differences in datasets, training/testing evaluation schemes, and segmented organs, we try to indicate how well our model performed with respect to recent state-of-the-art methods in Table \ref{tab:comparison}. In particualar, we provide a comparison to recent methods on two different datasets: (1) our own \textit{in-house} dataset for pancreas segmentation, acquired at Nagoya University Hospital, Japan, and consisting of 150 CT images; and (2) the publicly available \textit{TCIA Pancreas-CT} dataset of 82 patient images\footnote{\url{https://wiki.cancerimagingarchive.net/display/Public/Pancreas-CT} \citep{tciaPancreasCT} hosted by TCIA \citep{clark2013cancer}.} \citep{tciaPancreasCT}. For comparison with (2), we use the same 4-fold cross-validation (CV) split as in \citep{roth2015deeporgan,roth2017spatial}. 

Our results on dataset (1) achieves the highest reported performance in testing. On the other hand, our results on the public dataset (2) are comparable to other recent works that developed methods especially targeting this dataset and focusing on pancreas segmentation alone \citep{roth2017spatial,zhou2016pancreas}. 
\clearpage
\newpage
\begin{table}[ph!]
	\small
	\centering
	\caption{\textbf{Comparison to other methods.} We list other recent segmentation work performed on the same/similar datasets and organs and based on atlas-based segmentation propagation using 
		global affine \citep{wang2014geodesic}, 
		local non-rigid registration methods \citep{wolz2013automated} and in combination with machine learning (ML) \citep{tong2015discriminative}. We also list a method using regression forest (RF) and graph cut (GC) \citep{oda2016regression}, and two other methods utilizing 2D FCNs \citep{roth2017spatial,zhou2016pancreas}. Validation of other methods was performed using either leave-one-out-validation (LOOV) or cross-validation(CV). Best performance is shown in \textbf{bold}.}
	\label{tab:comparison}%
	\begin{tabularx}{\linewidth}{*{7}{X}}
		\hline
		\rowcolor[gray]{.9}\textbf{Method} & \textbf{Subjects} \ \ \ & \textbf{Approach}  \ \ \ \ \ & \textbf{Validation} \ \ \ \ & \textbf{Organs} \ \ \ \ \ \ & \textbf{Dice [\%]} \ \ \ \ \ \ \ \ \ & \textbf{Time [h]} \\
		\hline
		\hline
		&  &  &  &  &  &  \tabularnewline
		\multicolumn{7}{c}{\textbf{(1) In-house dataset}} \tabularnewline
		\hline		
		\textbf{Proposed} & 150   & 3D FCN & Testing & Liver & \textbf{95.4$\pm$ 2.0} & 0.07 \\
		&       &       &       & Spleen & \textbf{92.8$\pm$ 8.0} &  \\
		&       &       &       & Pancreas & \textbf{82.2$\pm$ 10.2} &  \\
		\rowcolor[gray]{.9}\citet{tong2015discriminative} & 150   & Global affine & LOOV & Liver & 94.9 $\pm$ 1.9 & 0.5 \\
		\rowcolor[gray]{.9}      &       &  + ML &       & Spleen & 92.5 $\pm$ 6.5 &  \\
		\rowcolor[gray]{.9}      &       &       &       & Pancreas & 71.1 $\pm$ 14.7 &  \\
		\citet{wang2014geodesic}& 100     & Global affine & LOOV & Liver & 94.5 $\pm$ 2.5 & 14 \\
		  &       &       &       & Spleen & 92.5 $\pm$ 8.4 &  \\
		&       &       &       & Pancreas & 65.5 $\pm$ 18.6 &  \\
		\rowcolor[gray]{.9}\citet{wolz2013automated}  & 150      & Local non-rigid & LOOV & Liver & 94.0 $\pm$ 2.8 & 3 \\
		\rowcolor[gray]{.9}      &       &       &       & Spleen & 92.0 $\pm$ 9.2 &  \\
		\rowcolor[gray]{.9}      &       &       &       & Pancreas & 69.6 $\pm$ 16.7 &  \\
		\citet{oda2016regression} & 147   & RF + GC & LOOV & Pancreas & 75.1 $\pm$ 15.4 & 3 \\
		\hline
		&  &  &  &  &  &  \tabularnewline
		\multicolumn{7}{c}{\textbf{(2) TCIA Pancreas-CT dataset}} \tabularnewline
		\hline
		\textbf{Proposed} & 82    & 3D FCN & 4-fold CV & Pancreas & 76.8 $\pm$ 9.4 & 0.07 \\ 
		\rowcolor[gray]{.9}\rowcolor[gray]{.9}\citet{roth2017spatial} & 82    & 2D FCN & 4-fold CV & Pancreas & 81.3 $\pm$ 6.3 & 0.05 \\
		\citet{zhou2016pancreas} & 82    & 2D FCN & 4-fold CV & Pancreas & \textbf{82.4$\pm$ 5.7} & n/a \\
		\hline
		&  &  &  &  &  &  \tabularnewline
	\end{tabularx}%
\end{table}%
\clearpage
\newpage 
\subsection{Direct comparison to 2D FCN networks} 
Furthermore, we implement the method of Zhou et al. \citep{zhou2016three, zhou2017deep} and apply it to the same dataset. This method employs a combination of three 2D FCNs trained on the orthogonal planes of the images. The results of each model are then fused by majority voting. This dataset consists of 240 3D CT scans with 18 manually annotated organs. A split of 228/12 cases was used for our training/testing as in \citep{zhou2016three, zhou2017deep}. A direct comparison can be seen in Table \ref{tab:2D_vs_3D}. It can be observed that our 3D FCN approach has a clear advantage for the smaller, thinner organs (like aorta, esophagus, gallbladder, inferior vena cava, portal vein, and prostate) but only performs comparable to the 2D FCNs when aiming at the larger organs (like lungs, liver, kidneys). Furthermore, a slightly higher overall performance can be observed for the average of all organ/vessel predictions when using the proposed cascaded 3D FCN approach.
\subsection{Computation} Training on 281 cases can take 2-3 days for 200-k iterations on a NVIDIA GeForce GTX TITAN X with 12 GB memory. However, in testing, the processing time for each volume was 1.4-3.3 minutes for each stage, depending on the size of the candidate regions; and 1.6-4.4 minutes using overlapping tiles in the second stage.
\textcolor{revision}{In order to achieve optimal GPU memory usage in training, we keep the input subvolume size at $N_\mathrm{x}\times N_\mathrm{y}\times N_\mathrm{z} = 132 \times 132 \times 116$, resulting in an output size of $44 \times 44 \times 28$ for each class output channel as in \citep{cciccek20163d}.}

\clearpage
\newpage
\begin{table}[htbp]
    \footnotesize
	\centering
	\parbox{15cm}{\caption{\small Direct comparison of the proposed cascaded 3D FCN approach against a 2D FCN approach using a majority voting scheme as in \citep{zhou2016three, zhou2017deep}. 18 different anatomical structures are compared using the Dice similarity score [\%]. Significantly better performance is shown in bold ($p<0.05$, Wilcoxon signed-rank test).}
	\label{tab:2D_vs_3D}}
	\sisetup{detect-weight=true,detect-inline-weight=math,separate-uncertainty}
	\begin{tabular}{l
			S[table-format=2.1,table-figures-uncertainty=1]
			S[table-format=2.1,table-figures-uncertainty=1]
			S[table-format=2.1,table-figures-uncertainty=1] 
			r}
		\hline
		\rowcolor[gray]{.9}\textbf{Label (Dice)} & \textbf{2D \citep{zhou2016three}} & \textbf{3D (stage1)} & \textbf{3D (stage2)} & \textbf{p-value}\tabularnewline
		\hline
		\textbf{right lung} 											& \bfseries 94.6\pm2.8	& 90.1\pm4.0 	& 91.9\pm3.6 			 & 0.028 \tabularnewline
		\rowcolor[gray]{.9}\textbf{left lung} 							& 92.8\pm3.8 			& 87.8\pm5.3	& 93.2\pm4.0 			 & 0.959 \tabularnewline
		\textbf{heart} 													& \bfseries  91.2\pm4.0 & 70.9\pm11.8 & 86.2\pm5.6 			 & 0.016 \tabularnewline
		\rowcolor[gray]{.9}\textbf{aorta} 								& 76.0\pm11.8 			& 50.7\pm5.4  & \bfseries  82.3\pm5.9  & 0.038 \tabularnewline
		\textbf{esophagus} 												& 24.6\pm16.7 			& 0.0\pm0.0   & \bfseries  51.9\pm5.3  & 0.011 \tabularnewline
		\rowcolor[gray]{.9}\textbf{liver} 								& \bfseries  94.3\pm3.3 & 90.2\pm3.5  & 93.6\pm2.7 			 & 0.049 \tabularnewline
		\textbf{gallbladder} 											& 47.5\pm39.9 			& 9.1\pm11.6  & \bfseries  58.4\pm33.2 & 0.011 \tabularnewline
		\rowcolor[gray]{.9}\textbf{stomach and duodenal} 				& 68.0\pm19.1 			& 58.2\pm15.2 & 61.9\pm13.4 			 & 0.070 \tabularnewline
		\textbf{stomach and duodenal (air)} 			 				& \bfseries  64.0\pm32.2& 52.4\pm27.4 & 48.8\pm26.6 & 0.001 \tabularnewline
		\rowcolor[gray]{.9}\textbf{stomach and duodenal (not air)}  	& 8.5\pm15.6 			& 1.1\pm1.7 	& \bfseries  20.7\pm20.0 & 0.000 \tabularnewline
		\textbf{spleen} 												& 86.7\pm14.5 			& 81.2\pm12.1 & 86.6\pm6.6 & 0.326 \tabularnewline
		\rowcolor[gray]{.9}\textbf{right kidney} 						& 92.2\pm2.1 			& 80.9\pm10.6 & 90.8\pm6.7 & 0.918 \tabularnewline
		\textbf{left kidney} 											& 90.2\pm4.0 			& 82.8\pm8.3 	& 86.1\pm11.1 & 0.179 \tabularnewline
		\rowcolor[gray]{.9}\textbf{inferior vena cava} 					& 63.6\pm19.1 			& 59.8\pm15.2 & \bfseries  70.6\pm17.5 & 0.007 \tabularnewline
		\textbf{portal vein} 											& 33.6\pm29.5 			& 30.0\pm21.3 & \bfseries  56.0\pm16.6 & 0.002 \tabularnewline
		\rowcolor[gray]{.9}\textbf{pancreas} 							& 55.4\pm19.3 			& 47.5\pm16.3 &  \bfseries 71.0\pm13.9 & 0.001 \tabularnewline
		\textbf{prostate} 												& 1.2\pm2.3 			& 0.0\pm0.0   & \bfseries  38.0\pm29.3 & 0.008 \tabularnewline
		\rowcolor[gray]{.9}\textbf{bladder} 							& 78.8\pm13.9 			& 57.2\pm14.0 & 71.6\pm20.8 & 0.213  \tabularnewline
		\hline
		\textbf{Mean}					    							& 65.0 & 52.8 & \bfseries 69.3 & 0.004 \tabularnewline
		\rowcolor[gray]{.9}\textbf{Std}   								& 33.4 & 32.2 & 26.1 &  \tabularnewline
		\textbf{Min} 						    						& 0.0  & 0.0  & 1.1   &   \tabularnewline
		\rowcolor[gray]{.9}\textbf{Max}									& 97.7 & 90.2 & 97.3 &  \tabularnewline
		\hline
	\end{tabular}
\end{table}%
\clearpage
\newpage
\clearpage
\newpage
\section{Fine-tuning to other datasets}
One advantage of deep learning based models is their ability to transfer learned features across dataset domains \citep{shin2016deep}. To this end, we trained a general FCN model employing the 3D U-Net architecture \citep{cciccek20163d} on the large dataset of CT scans including the major abdominal organ labels of Section \ref{sec:train-and-val}. This model can then be fine-tuned to other (smaller) datasets aiming at more detailed classification tasks or different field of views. For this purpose, we utilize separate training, fine-tuning, and testing datasets. As mentioned above, the general training set consists of 280 clinical CT images with seven abdominal structures (artery, vein, liver, spleen, stomach, gallbladder, and pancreas) labeled. 

\begin{figure}[htb]
	\centering	
	\adjincludegraphics[width=0.95\textwidth]{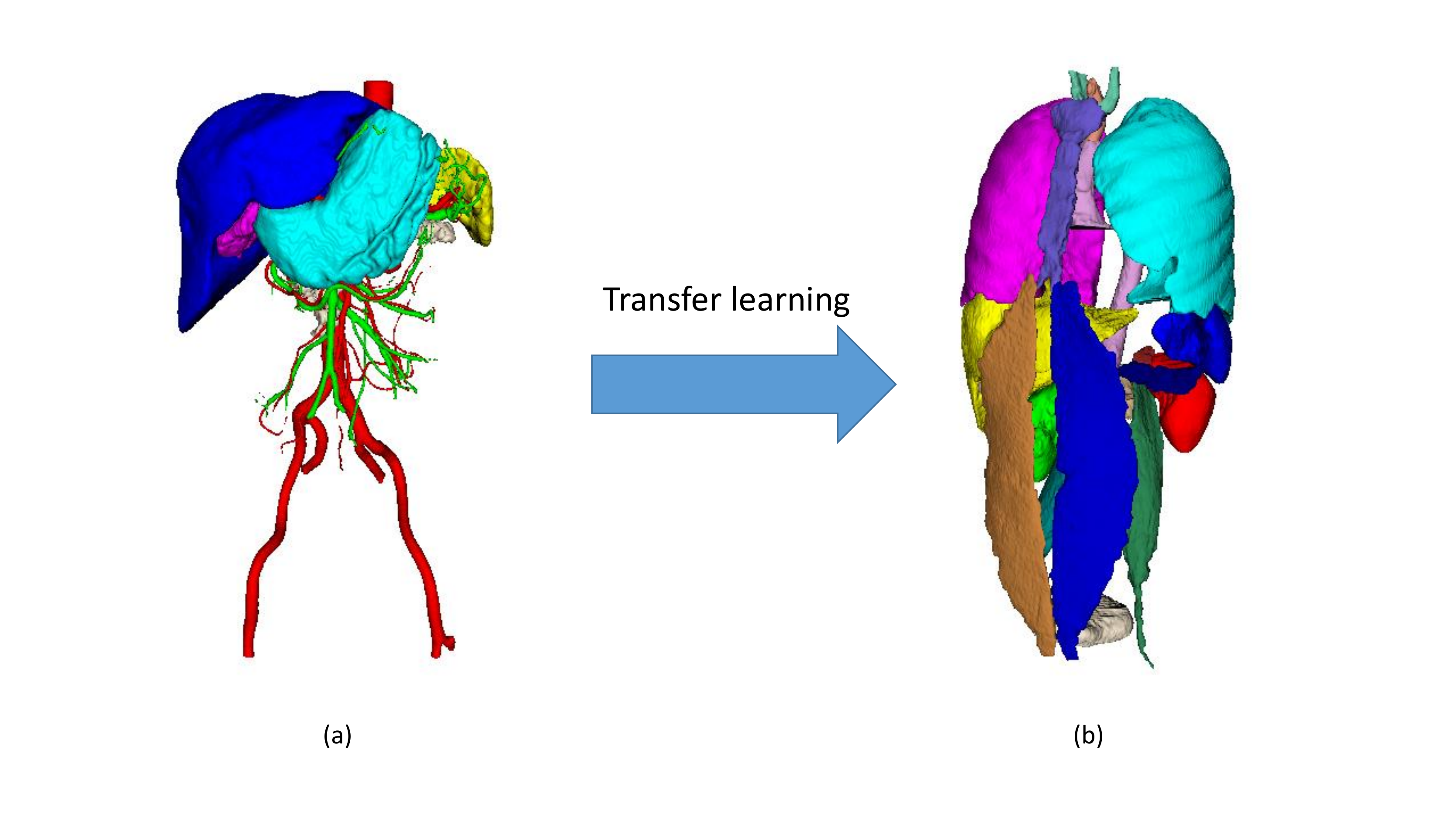}
	\caption{We fine-tune our model via transfer learning from 8 anatomical structures in the abdomen (a) to 20 anatomical structures in the whole torso (b). We show some typical ground truth labels that are used for training on both datasets.}
	\label{fig:fine-tuning}
\end{figure}

\begin{figure}[htb]
	\centering	
	\subfloat{\adjincludegraphics[valign=c,width=0.75\textwidth]{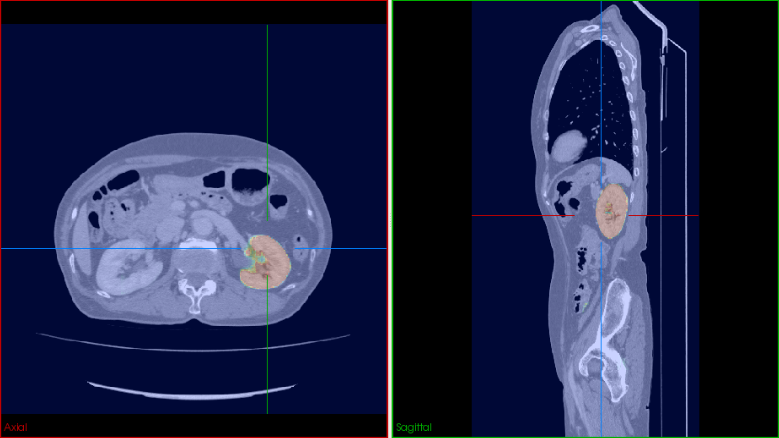}}
	\hspace{1.5em}
	\subfloat{\adjincludegraphics[valign=c,width=0.05\textwidth]{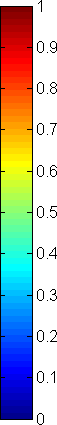}}
	\caption{Automated probability map for left kidney after transfer learning.}
	\label{fig:fine-tuning_kidney}
\end{figure}

We then fine-tune on a much smaller dataset consisting only of 20 contrast enhanced CT images from the Visceral Challenge dataset\footnote{\url{http://www.visceral.eu/benchmarks/anatomy3-open/}} \citep{jimenez2016cloud}, but with substantially more anatomical structures labeled in each image (20 in total). This fine-tuning process across different datasets is illustrated in Fig. \ref{fig:fine-tuning} with some ground truth label examples used for pre-training and fine-tuning. In fine-tuning, we use a 10 times smaller learning rate. We furthermore test our models on a completely unseen data collection of 10 torso CT images with 8 labels, including organs that were not labeled in the original abdominal dataset, e.g. the kidneys and lungs. A probabilistic output for kidney (not in the pre-training dataset) from our model is shown in Fig. \ref{fig:fine-tuning_kidney}.
\begin{figure}[htb]
	\centering	
	\adjincludegraphics[width=0.65\textwidth]{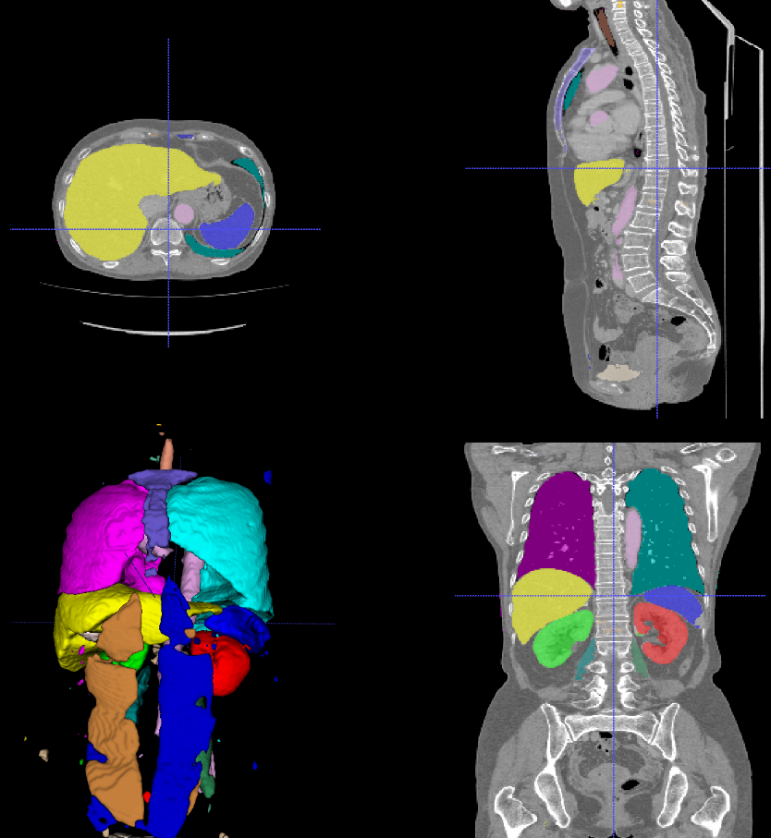}
	\caption{Multi-organ segmentation result. Each color represents an organ region on the unseen whole torso test set.}
	\label{fig:fine-tuning-result}
\end{figure}
\subsection{Fine-tuning results}
In testing, we deploy our fine-tuned model using a non-overlapping tiling approach as in previous sections. An automated segmentation result on the unseen test dataset by our fine-tuned model is shown in Fig. \ref{fig:fine-tuning-result}. Our fine-tuned approach provides a Dice score of right lung, left lung, liver, gall bladder, spleen, right kidney, left kidney, and pancreas are 0.96, 0.97, 0.95, 0.77, 0.90, 0.90, 0.88, and 0.36, respectively (summarized in Table \ref{tab:fine-tuning_dice_results_testing}). The relatively lower score for pancreas is due to several outlier cases on this dataset. These outliers are likely caused by variations of contrast enhancement across the datasets and the higher variability of the pancreas’ shape and intensity profile compared to other organs across different patients. 

Our approach and results, however, illustrate the generalizability and robustness of our models across different datasets. Fine-tuning can be useful when the amount of training examples for some target organs are limited. In this case, transfer learning achieves slight improvements over learning from scratch, especially in the kidneys (see Table \ref{tab:fine-tuning_dice_results_testing}). It should be noted that for this particular application, data augmentation already gives a good performance when learning models from scratch.
\begin{table}
	\footnotesize
	\centering
	\caption{\textbf{Testing on unseen whole torso dataset:} Dice scores [\%] for each segmented organ.}
	\label{tab:fine-tuning_dice_results_testing}
	\begin{tabular}{lrrrrrrrrr}
		\hline		
		\rowcolor[gray]{.9}\textbf{Dice} & \textbf{r. lung}	 & \textbf{l. lung}	 & \textbf{liver}	 & \textbf{gall}	 & \textbf{spleen}	 & \textbf{r. kidney}	 & \textbf{l. kidney}	 & \textbf{pancreas}	 & \textbf{Avg.}\tabularnewline
		\hline
		\textbf{scratch}	& 96.2 & 96.3 &	94.0 &	74.9 &	91.0 &	87.6 &	84.1 &	32.0 &	82.0 \tabularnewline
		\rowcolor[gray]{.9}\textbf{fine-tuned}	& 96.4 & 96.6 &	94.9 &	76.3 &	90.1 &	90.5 &	88.5 &	33.0 &	83.3 \tabularnewline
		\hline
	\end{tabular}
\end{table}

\section{Discussion}
The cascaded coarse-to-fine approach presented in this paper provides a simple yet effective method for employing 3D FCNs in medical imaging settings. No post-processing was applied to any of the FCN outputs. The improved performance stemming from our cascaded approach is especially visible in smaller, thinner organs, such as arteries and veins, particularly when compared to other recent FCN approaches using 2D FCNs \citep{zhou2016three}. Our results and recent literature indicate that 2D FCNs and especially the combination of orthogonally applied 2D FCNs \citep{zhou2016three,roth2017spatial,zhou2016pancreas} might be sufficient for larger and mid-sized organs. In fact, the combination of 2D FCNs even slightly outperforms our 3D approach for some organs. On the other hand 3D convolutional kernels are important for distinguishing the thin (vessel-like) and small organs as can be seen in the improved performance of our approach. When compared to other cascaded approaches using 2D FCNs that focus on single organs \citep{roth2017spatial,zhou2016pancreas}, we perform similar to the state of the art. Our findings are also consistent with \citep{roth2017spatial,zhou2016pancreas} that show that cascaded approaches are useful for applying deep learning methods to medical image segmentation. Note that we used different datasets (from different hospitals and scanners) for separate training/validation and testing. These experiments illustrate our method\rq{}s generalizability and robustness to differences in image quality and populations. Running the algorithms at half resolution allows efficient training on a single GPU. In contrast, using the same field of view for each subvolume with the original resolution would require $8\times$ more memory with the current architecture and would force us to reduce the amount of context visible to the 3D FCNs. In this work, we utilized 3D U-Net for the segmentation of CT scans. However, the proposed cascaded approach in principle should also work well for other 3D CNN/FCN architectures and 3D image modalities. Exploration of other loss functions such as the Dice score \citep{milletari2016v,li2017compactness} could help further in dealing with the class imbalance issue. We used Caffe's stochastic gradient descent solver \citep{jia2014caffe} for all experiments in this work. Alternative optimizers could further improve training performance \citep{kingma2014adam}. 

In the future, prediction results from different models could be combined in order to achieve the best overall performance. Furthermore, additional anatomical constraints could be included in order to guarantee topologically correct segmentation results \citep{bentaieb2016topology,oktay2017anatomically}. With growing amounts of available GPU memory, the need for computing overlapping sub-volume predictions as in this work will be reduced as it will be come possible to reshape the network to accept arbitrary 3D input image sizes \citep{long2015fully}.
\section{Conclusion}
In conclusion, we showed that a cascaded deployment of volumetric fully convolutional networks (3D U-Net) can produce competitive results for medical image segmentation on a clinical CT dataset while being efficiently deployed on a single GPU. An overlapping tiles approach during testing produces better results with only moderate additional computational cost. The proposed method compares favorably to recent state-of-the-art work on a completely unseen dataset. Our results indicate that 3D convolutional features are advantageous for detecting smaller organs and vessel. A promising future direction might be hybrid approaches that combine 2D and 3D FCN-type architectures at multiple scales. We have made our code, pre-trained models, and fine-tuned models available for download\footnote{\url{https://github.com/holgerroth/3Dunet_abdomen_cascade}} in order to allow further applications and fine-tuning to different datasets.
\linebreak
\linebreak
\textbf{Acknowledgments } This paper was supported by MEXT KAKENHI (26108006, 26560255, 25242047, 17H00867, 15H01116) and the JPSP International Bilateral Collaboration Grant. 
\begin{flushleft}
\vspace{2em}
\textbf{Conflict of interest statement: } The authors declare that they have no conflict of interest.
\end{flushleft}
\section*{References}
\bibliography{cmig2017_rothhr_cascaded_fcn}

\end{document}